\documentclass[runningheads]{llncs}
\usepackage{graphicx}
\usepackage{hyperref}

\usepackage{svg} 
\usepackage{algorithm} 
\usepackage{algorithmic} 
\usepackage{color} 
\usepackage{array} 

\makeatletter
\newcommand{\printfnsymbol}[1]{%
  \textsuperscript{\@fnsymbol{#1}}%
}
\makeatother

\definecolor{HSafron}{RGB}{232,125,30}

\begin{document}
\title{Improving Image Classification Robustness through Selective CNN-Filters Fine-Tuning}
%
\titlerunning{Selective CNN-Filters Fine-Tuning}
%
\author{Alessandro Bianchi\thanks{equal contribution}
\inst{1} \and
Moreno R. Vendra\printfnsymbol{1}
\inst{1}  \and \\
Pavlos Protopapas\inst{2} \and
Marco Brambilla\inst{1}}
\authorrunning{M. Vendra et al.}
%
\institute{Politecnico di Milano, Milan 20133, Italy \and
Harvard University, Cambridge MA 02138, USA}
\maketitle              
\begin{abstract}
Image quality plays a big role in CNN-based image classification performance. Fine-tuning the network with distorted samples may be too costly for large networks. To solve this issue, we propose a transfer learning approach optimized to keep into account that in each layer of a CNN some filters are more susceptible to image distortion than others. Our method identifies the most susceptible filters and applies retraining only to the filters that show the highest activation maps distance between clean and distorted images. Filters are ranked using the Borda count election method and then only the most affected filters are fine-tuned. This significantly reduces the number of parameters to retrain. We evaluate this approach on the CIFAR-10 and CIFAR-100 datasets, testing it on two different models and two different types of distortion. Results show that the proposed transfer learning technique recovers most of the lost performance due to input data distortion, at a considerably faster pace with respect to existing methods, thanks to the reduced number of parameters to fine-tune. When few noisy samples are provided for training, our filter-level fine tuning performs particularly well, also outperforming state of the art layer-level transfer learning approaches.
\end{abstract}

\section{Introduction} \label{introduction}

In recent years, deep neural networks (DNNs) have become increasingly good at learning a very accurate mapping from inputs to outputs, from large amounts of labeled data, across different applications \cite{DBLP:journals/corr/JaderbergSVZ14a,Karpathy:2017:DVA:3069214.3069250,DBLP:journals/corr/MnihKSGAWR13,DBLP:journals/corr/OordDZSVGKSK16}.

The unfortunate downside of these models is their inability to generalize whenever they are presented different conditions from the ones that they have encountered during training. In the literature, this scenario is known by the name of \textit{dataset shift} \cite{Moreno-Torres/2012/UVD/2030819/2031236}. More formally, in classification problems, where the need for the model is to learn a mapping from \(X\rightarrow Y\), the class label $y$ is causally determined by the values of the covariates $x$, which together determine the joint distribution $P(x,y)$. \textit{Dataset shift} appears whenever training and test joint distributions ($P_{tr}(x,y)$ and $P_{tst}(x,y)$, respectively) are different. That is, when $P_{tr}(x,y) \ne P_{tst}(x,y)$.

The aforementioned deep learning models are commonly trained on carefully annotated datasets. In most realistic applications, data is collected from an infinite number of novel and cluttered scenarios, many of which would not be represented by the properly annotated data on which the model was trained on. The trained model would then be ill-prepared to make predictions on this new type of "noisy" data.

Obtaining accurately annotated data is a tedious process and often impractical in many situations. It would then be ideal to transfer knowledge gathered by the model on undistorted data (\textit{source domain}), to enable training of another model to properly classify noisy inputs (\textit{target domain}), even when few labeled observations from this noisy setting are available. 

In our work, we focus on applying this \textit{transfer learning} approach to the context of image classification, where, starting from a model trained on undistorted images, we assess which are the parts of the network that are most sensitive to input data distortion, fine-tuning them with noisy inputs such that the model is then able to properly classify distorted images.

In most realistic computer vision applications, an input image undergoes some form of image distortion including blur and additive noise during image acquisition, transmission or storage. However, most popular large scale datasets do not have images with these artifacts, thus testing pre-trained DNN models (trained on such images) on slightly perturbed ones, would cause a considerable degradation in classification performance, even though the added distortion does not hinder the human ability to classify the same images \cite{DBLP:journals/corr/DodgeK16}.

Borkar and Karam \cite{DBLP:journals/corr/BorkarK17} proved that features learned from a dataset of high quality images are not invariant to image distortion or noise and cannot be directly used for applications where the quality of images is different than that of the training images. The feature spaces of the \textit{source} and \textit{target domain} are different (i.e., $X_S \neq X_T$).

Through our study, we try to address this problem directly acting on on the features learned by the model on the \textit{source domain}, so that they become invariant to image distortion, ideally obtaining features such that $X_S \approx X_T$.

As in \cite{DBLP:journals/corr/BorkarK17}, we prove that among all the filters of the convolutional layers of a DNN, some filters are more susceptible to input distortions than others and that correcting the activations of these filters can help recover the lost performance. However, instead of correcting the activations, like in \cite{DBLP:journals/corr/BorkarK17}, we act directly on the filters, so that the learned features become nearly invariant to image distortion.

Leveraging this finding, \textbf{we propose a novel technique to rank the filters that are most affected by input data distortion, through an appropriate distance metric and voting technique} that we will detail in Section \ref{noise-impact-analysis}.

We will then present a detailed methodology to directly act on a subset of the aforementioned most affected filters per layer, so that the output activations become robust to image distortion, preventing the considerable degradation in classification performance that would normally take place if the DNN was only trained on undistorted inputs.

Differently from the usual fine-tuning of pre-trained models, that rely on retraining all the filters in some convolutional layers of the network, we show that fine-tuning only a subset of the most affected filters of those layers, achieves a better overall performance and a lower training cost than retraining all the filters of those layer, when data in the \textit{target domain} is limited.

The remainder of the paper is organized as follows. Section \ref{problem-definition} details a more formal definition of the problem. Section \ref{related-work} provides an overview of the related work in assessing and improving robustness of DNNs to input image perturbations. Section \ref{background} clarifies some of the necessary tools and ideas described in this paper. Section \ref{methodology} presents a detailed description of our proposed approach, while Section \ref{experimental-setting} delineates the experiments to validate the technique, whose results will be discussed in Section \ref{results-and-discussion}. Concluding remarks are given in Section \ref{conclusion}.

\subsection{Problem} \label{problem-definition}
\subsubsection{Problem Setting}
To provide a more formal definition of the problem, suppose we have a dataset $D_{T}$, of limited size, where $D_{T} = \{(x_{1}^{T}, y_{1}), ..., (x_{n}^{T}, y_{n})\} \subseteq \mathcal{X_T} \times \mathcal{Y}$, in which $\mathcal{X_T}$ and $\mathcal{Y}$ respectively denote the domain of predictors $X_T$ and classes $Y$, while $n = |D_{T}|$. We know for a fact that this set of collected data points have undergone some form of image distortion, which can be modeled as a function $g(\cdot)$. $D_{T}$ can then be described as $D_{T} = \{(x_{1}^{T}, y_{1}), ..., (x_{n}^{T}, y_{n})\} = \{(g(x_{1}^{C}), y_{1}), ..., (g(x_{n}^{C}), y_{n})\}$ where $x_{i}^{C}$ represents the undistorted data point, sampled from a "clean" dataset $D_{C} = \{(x_{1}^{C}, y_{1}), ..., (x_{n}^{C}, y_{n})\} \subseteq \mathcal{X_C} \times \mathcal{Y}$, in which $\mathcal{X_C}$ and $\mathcal{Y}$ respectively denote the domain of predictors $X_C$ and classes $Y.$

The goal is to use the set of distorted data points $D_{T}$ to train a DNN to perform a generic classification task. If $D_{T}$ is not of sufficient size to train a DNN model from scratch (with random initialization), we may have to move to pre-trained models. Focusing our attention to image classification settings, a pre-trained model is a convolutional neural network (CNN) which has been trained on a very large set of undistorted images $D_{S} = \{(x_{1}^{S}, y_{1}^{S}), ..., (x_{m}^{S}, y_{n}^{S})\}$ $ \subseteq \mathcal{X_S} \times \mathcal{Y_S}$ (where $\mathcal{X_S}$ and $\mathcal{Y_S}$ respectively denote the domain of predictors $X_S$ and classes $Y_S$, while $m = |D_{S}|$) which is known to perform very well in the task of classifying images. 

Nevertheless, as shown in \cite{DBLP:journals/corr/DodgeK16}, testing distorted images with a pre-trained DNN model, even though such image distortions $g(\cdot)$ do not represent adversarial samples for the DNN, results in a considerable drop in classification performance. The reason for this degradation is attributed to the distortion function $g(\cdot)$, which is responsible for increasing the effects of the phenomenon known as \textit{covariate shift} \cite{Moreno-Torres/2012/UVD/2030819/2031236}, defined as the case in which there is a change in the distribution of the input variables between training and testing data. More formally, assuming that the distribution of the undistorted images $P(X_C)$ is the same distribution of the images used to train the pre-trained model, we have that $P(X_C) = P(X_S)$. However, because of the distortion function $g(\cdot)$, the distribution of the "clean" images, is different from the distribution of their distorted counterparts (i.e., $P(X_C) \neq P(X_T)$), which means, due to the transitive property, that $P(X_C) \neq P(X_S)$. This result indicates that features learned from a dataset of high quality images, such as $D_S$, are not invariant to image distortion or noise, and cannot be directly used for applications where the quality of images is different than that of the training images ($D_T$).

\subsubsection{Problem Definition}
The goal for our work is to show an approach to potentially shrink the effect of the \textit{covariate shift} caused by the distortion function $g(\cdot)$, to eventually come up with a model that is robust to this type of phenomenon, leveraging the potential of the proposed \textit{transfer learning} method, while keeping the number of parameters to train into a feasible range, considering the limited amount of training samples. Starting from a CNN trained on pristine images, we demonstrate that in each of its convolutional layers, some filters are more susceptible to image distortion than others. Through a proper election method, we rank these filters based on the impact that distortion has on them. Finally, we fine-tune only the most affected ones, eventually achieving the goal of learning convolutional filters that are nearly invariant to input data distortion.

\subsection{Related Work} \label{related-work}
In literature we can find a multitude of approaches that address the \textit{covariate shift} issue, in image datasets, due to input data distortion. Starting from \cite{DBLP:journals/corr/DodgeK16} for example, which determined how different kinds and intensities of image distortion affect CNNs performance on image classification; they found all networks they tested to be susceptible both to blur and noise, but they observed that deeper networks performance falls off slower than the one of shallower networks; they concluded that deeper structures give the network more room to learn features that are not affected by noise. They also observed that blur does not significantly affect early filter responses, but in spite of this the last layer activations exhibit significant changes, so even slight differences in early layers propagate to deeper layers. Noise on the other hand causes many activations in the first layer due to its high frequency nature, and this translates in significant changes in the last layer responses.

\cite{DBLP:journals/corr/ZhouSC17} performed a similar analysis on the effects of distortions on image classification, but they also proceeded to propose two solutions to such problem: fine-tuning and retraining. In fine-tuning they start from a pre-trained model and continue training the first N layers with distorted samples while keeping the rest of the network fixed; when performing retraining instead they train the entire network on the distorted dataset starting from random weights. They observed that both techniques reduce the classification error rate of distorted samples, but such improvement in performance depends on the network and training dataset size: if the number of trainable parameters is large, fine-tuning is a better alternative than retraining since it is less prone to overfitting on small datasets. They also show that both fine-tuning and retraining tend to “adjust” the image representation, making it similar to the representation of the undistorted image from the pre-trained model; to prove this they show that, for blurred images, fine-tuning and retraining both increase the variance of the gradient of the activations, showing that such activations actually contain more information with respect to the ones produced by the original model on the distorted images.

Following the work of \cite{DBLP:journals/corr/ZhouSC17}, \cite{DBLP:journals/corr/BorkarK17} proposed an alternative solution to improve the performance of pre-trained CNNs on distorted images: the authors observed that for each layer of a CNN, certain filters are far more susceptible to input distortions than others and that correcting the activations of these filters can help recover lost performance. Starting from this observation they rank the convolutional filters in order of the highest gain in classification accuracy upon correction; then they proceed to correct the activations of the top ranked filters by appending small blocks of stacked convolutional layers at their outputs, and training them while keeping the rest of the network fixed. By doing so, they are able to significantly improve the robustness of the network against image distortions while reducing the number of trainable parameters and achieving faster convergence in training with respect to fine-tuning entire layers. That said, the amount of parameters to train to implement this technique remains considerably high with respect to our approach; moreover, the ranking of the filters is applicable only when the clean and noisy version of the same image is available, and this may not always be a viable option in real world problems. 

\section{Background} \label{background}
\subsection{Convolutional Neural Networks} \label{convolutional-neural-network}
Here we provide a brief overview of deep neural networks based on \cite{Goodfellow-et-al-2016}. A deep neural network is a collection of small and simple elements called neurons, organized in multiple layers, thus the adjective \textit{deep}. In general each neuron computes the following function:
\begin{equation}
f(\textbf{x}) = \phi(\textbf{w}^T\textbf{x}+b)
\end{equation}
where \textbf{x} is the input, \textbf{w} is a weight vector, \textit{b} is a bias term and $\phi$ is a nonlinear function.
The nonlinear function is an important aspect of the neuron composition because it allows each layer to learn a non-linear function of its inputs; since the network is composed of stacked layers, the output of each layer becomes the input of the following one.

For image recognition problems, the network receives in input an image; usually each input neuron receives information about the entire input sample, but in this case connecting all neurons to all pixels in the image would result in prohibitive memory and computation requirements. To solve this issue we connect each neuron to a small, locally connected portion of the input called \textit{receptive field}, and we move this window across all the image; this way the number of parameters and the quantity of memory needed to compute each sample is drastically reduced. Such technique, also known as \textit{weight sharing}, is equivalent to convolutional filtering where the filters are represented by the shared weight vectors. Layers arranged in this way are called \textit{convolutional layers}.

Following convolutional layers we usually find pooling layers; such layers apply a pooling operation over a window of fixed size across the convolutional layer response, producing in output a single value for each input region. There are different kinds of pooling operations, but the most common one is the max operator, where the maximum neuron response in the window is produced as output. These layers serve two purposes: they improve noise robustness and increase the size of the \textit{receptive field} in deeper layers without increasing the size of the filters.

The last stage of the network is usually a softmax layer, which translates the output of the network to probabilities over the output classes. The output of this layer is then compared to the true labels at training time and a loss function representing how closely the predictions match the true labels is computed. The gradient of this cost function is then propagated backward through the network to compute the updates for the parameters of each neuron.

\subsection{Image distortion} \label{image-distortion}
Deep neural networks structured as described before, also known as \textit{convolutional neural networks}, have been able to deal with multiple complex computer vision tasks, achieving particularly good results in image classification. The availability of large high quality image datasets \cite{5206848} \cite{Everingham2010} has been crucial for successfully training very deep and complex networks. 

In practical applications though it is common for images to be affected by different kinds of distortions, usually in the form of blur or noise. Blur can be caused by camera shake or lack of camera focus and it can also affect pictures taken with high quality cameras; noise on the other hand is usually due to bad lighting conditions or high sensor temperatures \cite{DBLP:journals/corr/ZhouSC17}. Distortions like these may impact the ability of many deep learning models to perform as well as they do when they are presented with clean images: \cite{DBLP:journals/corr/VasiljevicCS16} shows how state-of-the-art models trained on high-quality image datasets make unreliable, albeit low-confidence, predictions when they encounter blur in their inputs, due to their inability to generalize from their sharp training sets; \cite{DBLP:journals/corr/DodgeK16} instead demonstrates that both noise and blur cause significant differences in the outputs of convolutional layers, when comparing clean and distorted images. In both cases of distortion though a technique called \textit{fine-tuning} was found capable of recovering most of the lost performance; it will be described in the next paragraph \cite{DBLP:journals/corr/VasiljevicCS16} \cite{DBLP:journals/corr/ZhouSC17}.

\subsection{Transfer Learning} \label{transfer-learning}
As defined in \cite{5288526}:

\begin{center}
Given a \textit{source domain} D\textsubscript{S} and learning task T\textsubscript{S}, a \textit{target domain} D\textsubscript{T} and learning task T\textsubscript{T}, \textit{transfer learning} aims to help improve the learning of the target predictive function f\textsubscript{T}($\cdot$) in D\textsubscript{T} using the knowledge in D\textsubscript{S} and T\textsubscript{S}, where D\textsubscript{S}$\neq$D\textsubscript{T}, or T\textsubscript{S}$\neq$T\textsubscript{T}.
\end{center}

Many applications of \textit{transfer learning} techniques have been studied \cite{5288526}; in practice, in most deep learning scenarios doing \textit{transfer learning} means transferring weights from an existing network trained on a large dataset to another network. Training a network with limited data, poor initialization, and a lack of regularization to control capacity, may very easily and slowly lead to find a sub-optimal minima, thus failing to learn how to generalize well \cite{DBLP:journals/corr/NeyshaburBMS17}. This is why reusing the weights of another network as initialization can boost performance even when small target training sets are available. In particular in \cite{DBLP:journals/corr/YosinskiCBL14} \cite{DBLP:journals/corr/ZeilerF13} it was observed how the first layers of the network tend to learn simpler features that generalize more easily to different tasks, with respect to deeper layers; for networks trained on similar tasks and datasets, freezing the first few layers of the source network, and training the remaining layers at a low learning rate proves to be a good strategy. This process is called fine-tuning.  

\section{Methodology} \label{methodology}

This section is devoted to present in detail the steps of our proposed methodology, which can be summarized through the following points: 1) We train a CNN on a source dataset of pristine images, to serve as our baseline model; 2) We measure the susceptibility of convolutional filters to input distortion using a peculiar distance metric and rank said filters in the order of highest susceptibility to input distortion; 3) We fine-tune only a subset of these most affected filters, with a target dataset of distorted images, to attain a significant improvement in terms of robustness of the network against input data distortion, achieved at a notably smaller computational cost than the usual layer-wise fine-tuning techniques. 


\subsection{Model training on source dataset} \label{model-training-on-source-dataset}

In order to simulate an authentic \textit{transfer learning} scenario, first, we need a model to be used as a starting point. Adopting the same notation of Section \ref{problem-definition}, let's define $M$ as a CNN, trained on a \textit{source dataset} $D_S$ of undistorted images, to carry out the task of image classification (i.e., learning a mapping from an input image $x_{i}^{S}$ to its corresponding label $y_{i}^{S})$. Since this baseline model $M$ has only been trained to classify images in a "clean" data scenario, the model will not perform as well when trained on "noisy" data, as extensively proven in \cite{DBLP:journals/corr/DodgeK16}.

A fundamental aspect of our work is then to show how to assess which are the parts of the network $M$ that are most affected by the difference between the "clean" input data distribution (\textit{source domain}) and the "noisy" one (\textit{target domain}). This is what will be detailed in the upcoming section.

\subsection{Noise impact analysis} \label{noise-impact-analysis}

\begin{figure}[t!]
    \centering
    \includegraphics[scale=0.5]{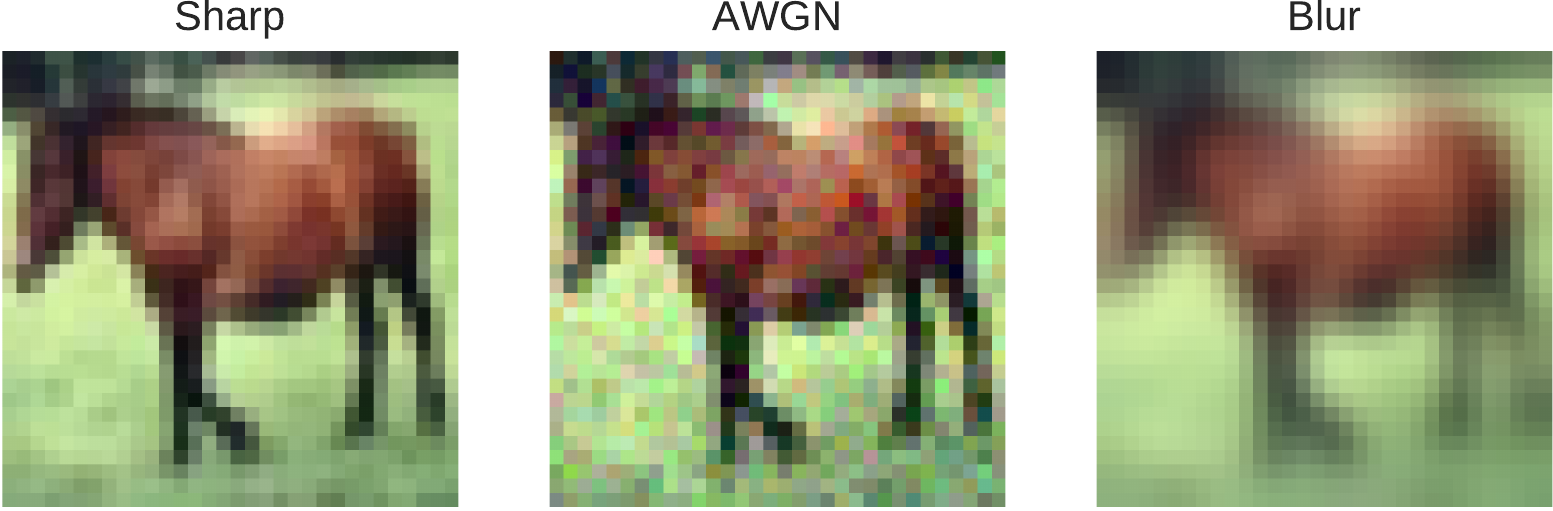}
    \caption{Examples of the \textit{associative} case: sharp and distorted images (with Additive White Gaussian Noise and blurring, respectively) refer to the same sample.}
    \label{fig:image-pairs-examples}
\end{figure}

Since our study addresses the problem of image classification, we focused our attention to the portion of the network that is responsible to learn the latent representation of every input image. To this regard, our study addresses the problem of finding a quantitative measure to determine the change introduced by noise, with respect to the clean setting, at the convolutional layers level. Being the convolutional layer the core building block to learn the latent representation of every input image, the comparison is performed at the 2-dimensional activation maps level, as we are willing to evaluate how noise impacts the learned latent representation of every image.

To compute this difference, a \textit{(clean, noisy)} image pair is necessary to perform the comparison. For empirical reasons, it is crucial to make a substantial splitting, depending on the setting in which we want to take advantage of the proposed technique: we will refer to \textit{associative pairs} whenever the \textit{(clean, noisy)} image pair concerns the same image. In this scenario, adopting the same notation of Section \ref{problem-definition}, being $x_{i}^{C} \in D_C$ the clean image, and $x_{i}^{T} \in D_T$ the noisy one, then $x_{i}^{T} = g(x_{i}^{C})$, where $g(\cdot)$ is the function for the input distortion. Examples of images that display this scenario can be observed in Figure \ref{fig:image-pairs-examples}.

We will instead refer to \textit{non associative pairs} whenever this \textit{(clean, noisy)} image pair will not refer to the same image ($x_{i}^{T} \neq g(x_{i}^{C})$), to accommodate circumstances in which the clean and noisy pair of the same image will not be available.

Being the \textit{non associative} case a slight variation of the \textit{associative} one, we present the methodology assuming an \textit{associative} setting. Details on the \textit{non associative} scenario will be given in Section \ref{non-associative-method}. 

\subsubsection{Measuring filters' susceptibility to input data distortion} \label{measuring-filters-susceptibility}

\begin{figure}[t!]
    \centering
    \includegraphics[width=\textwidth]{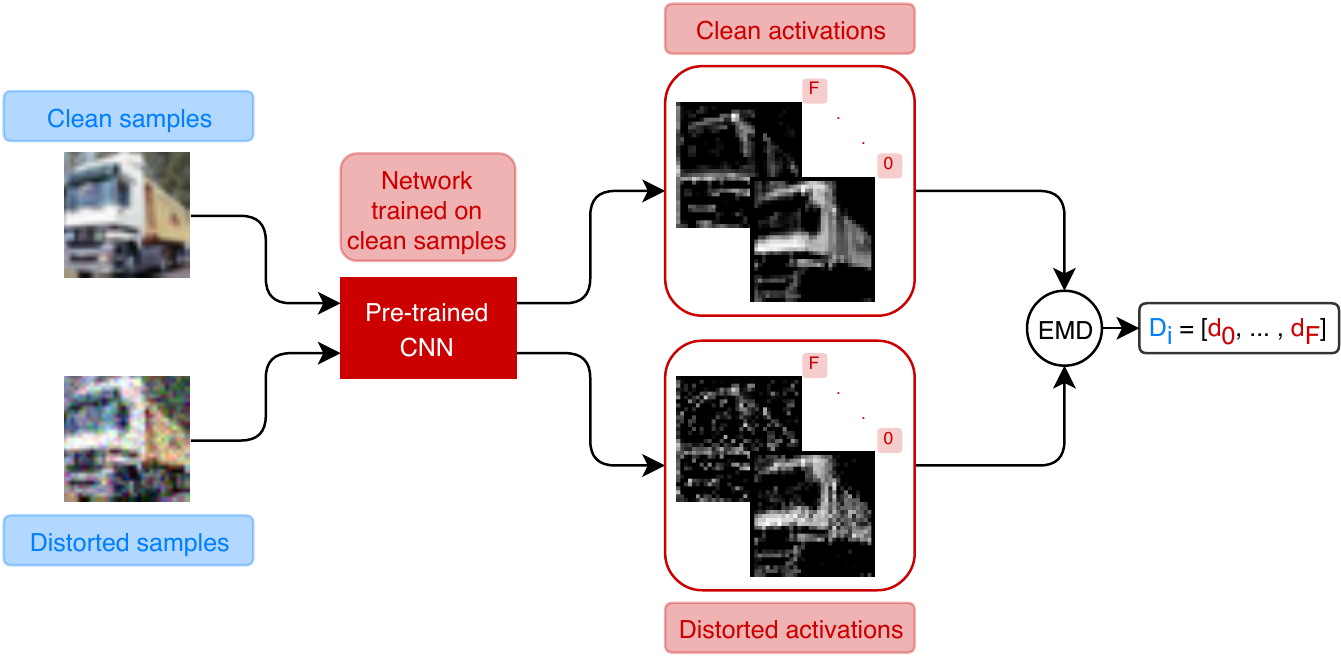}
    \caption{High level diagram of the \textit{associative} method. The clean and noisy version of the same image are fed into the same baseline model, producing two different sets of feature maps activations, which are then compared using the \textit{EMD} metric to generate an array of distances, where each element represents the scalar distance between each of the single feature map activations in the investigated layer.}
    \label{fig:high-level-methodology}
\end{figure}

The fundamental objective of the noise impact analysis is to measure which are the feature maps, for each convolutional layer, that are most affected by input data distortion.
Simply put, the goal is to measure which are the feature maps that vary the most between their clean and noisy counterparts. This variation is a measure of how much each convolutional filter, which generated each corresponding feature map activation, is sensitive to input data distortion, indicating how much each filter is responsible for the performance drop of the baseline model when tested on noisy inputs.

To measure this variation between feature maps activations, because of its informal intuition, we adopted the \textit{Earth mover's distance} (\textit{EMD}), which is also known, in mathematics, as the Wasserstein metric. In principle, if the distributions - that in our case are represented as feature maps activations - are treated as two different ways of piling up a certain amount of dirt over a certain region $R$, then the \textit{EMD} represents the minimum cost of turning one pile into the other, where the cost is assumed to be the amount of dirt moved, times the distance by which it was moved \cite{Rubner:1998:MDA:938978.939133}.

On a purely iterative fashion, repeating the comparison for each \textit{($x_{i}^{C}$, $x_{i}^{T}$)} pair of training images, we can now compute, for each convolutional layer, the \textit{EMD} metric between the feature map activations generated when the clean version of the image is fed into the baseline model $M$, and the feature maps activations generated when the noisy version of the same image is fed into the network. An high level diagram of the detailed approach is shown in Figure \ref{fig:high-level-methodology}.

After having computed the \textit{EMD} for each \textit{($x_{i}^{C}$, $x_{i}^{T}$)} input pair, we now have a rectangular distance matrix $D$, of shape $N$ x $F$, where $N$ represents the size of the training dataset ($N = |D_C| = |D_T|$), while $F$ describes the number of feature maps activations present in the examined convolutional layer $C$. This matrix now represents, for each image at index $i \in \{1, ..., N\}$, how distant each feature map activation at index $j \in \{1, ..., F\}$ is between its clean and noisy form. 

\subsubsection{Ranking filters by susceptibility to input distortion} \label{ranking-filters}
The remaining aspect of the noise impact analysis consists in aggregating all the computed distances, to determine which are the feature maps that are most affected by noise. Associating a large value of \textit{EMD} to a greater change due to noise, the objective is to produce a final ranking that sorts, from the most changing feature map, to the least changing one, all the feature maps indices in the examined convolutional layer $C$. To accomplish  this task, we relied on an election method known by the name of \textit{Borda count} \cite{MathInSociety}. The voters (images) rank the list of candidates (feature maps) in order of preference. Points are then given to each candidate in reverse proportion to their ranking, so that higher-ranked candidates receive more points. When all votes have been counted, and the points added up, the candidate with the most points wins. In our setting, candidates receive 10 points each time they are ranked first, 9 for being ranked second, and so on, with the 10\textsuperscript{th} candidate receiving just 1 point, while the remaining ones receive 0 points. A general overview of the pseudo-code related to this approach can be appreciated in Algorithm \ref{alg:borda}.

\begin{algorithm}[tb]
  \caption{Ranking convolutional filters per layer with Borda count}
  \label{alg:borda}
\begin{algorithmic}
  \STATE {\bfseries Input:} convolutional layer $l$, input images $samples$
  \FOR{$filter$ {\bfseries in} $l$}
  \STATE Initialize $borda\_count$ of $filter$ = 0
  \ENDFOR
  \FOR{$(clean\_img, noisy\_img)$ {\bfseries in} $samples$}
  \STATE $distance$ = {\bfseries compute\_distance}($l$, $(clean\_img, noisy\_img)$) 
  \STATE $ranked\_filters$ = {\bfseries argsort}($distance$)
  \STATE Truncate $ranked\_filters$ at 10 items
  \STATE Initialize $idx$ = 0
  \WHILE{$idx < 10$}
  \STATE $candidate\_filter$ = $ranked\_filters$ at position $idx$
  \STATE $points$ = 10 - $idx$
  \STATE Increase $borda\_count$ of $candidate\_filter$ by $points$
  \STATE Increase $idx$ by 1
  \ENDWHILE
  \ENDFOR
\end{algorithmic}
\end{algorithm}


Because this voting technique tends to elect broadly-acceptable candidates, rather than those preferred by a majority (i.e. only the one that varies the most, the majority of the times), \textit{Borda count} proved to be the most successful technique to provide the ranking for the feature maps most affected by noise.

\subsection{Model fine-tuning on target dataset} \label{model-fine-tuning-on-target-dataset}

Once we assessed which are the feature maps activations that change the most due to the input noise, we can proceed to fine-tune the filters that produced them. Parameters fine-tuning in convolutional neural networks is usually performed taking a previously trained network, and retraining some of its convolutional layers, in order to adapt them to the target task or domain. Since our objective is not to accomplish a new task, but rather to mitigate the noise impact on the source model, we are going to retrain only the filters that produced the most changing activation maps, instead of all the filters in the layer.

In details, we will train a "\textit{partially trainable}" neural network, where, unlike the standard \textit{transfer learning} approach of "freezing" at the layer level (i.e. keeping constant all the parameters for the layer), we will do it at the filter level. As a result, within the same convolutional layer $C$, some filters will be re-trainable, while others will not. 

The intuition behind this approach is that, by only acting on the filters that are most affected by noise (i.e. the most corrupted ones), we directly act on the parts of the network that are most responsible for the performance drop when moving from a clean data setting to a noisy one. 



\subsection{Non associative method} \label{non-associative-method}

In Section \ref{noise-impact-analysis}, we assumed to have a \textit{(clean, noisy)} input pair of the same image. In practical applications, this might not always be the case. In the previously defined approach, not having the \textit{(clean, noisy)} input pair for the same image, would prevent the applicability of the technique, as it would not be possible to compute the \textit{EMD} between feature maps of the clean and noisy image pair. In this \textit{non associative} setting, each noisy image $x_{i}^{T}$ is not generated applying the distortion function $g(\cdot)$ to its corresponding clean version $x_{i}^{C}$. Instead, it is an independently collected noisy image, of which its corresponding clean version is not available. Formally, $x_{i}^{T} \neq g(x_{i}^{C})$.

To overcome this problem, we had to rely on the idea of finding images that would serve as representatives for a given class of images, where the class labels, in this case, would not be their actual categories. Instead, images would be split based on whether they were noisy or not. The two representatives would then be compared against each other to compute the \textit{Earth mover's distance}. Inspired by these ideas, we decided to rely on the concept of clusters \textit{exemplars}.

\subsubsection{Exemplars} \label{exemplars}
The \textit{k-medoids} algorithm is a clustering algorithm that breaks the dataset up into groups, attempting to minimize the distance between points labeled to be in a cluster and a point designated as the center of that cluster \cite{kaufman1987clustering}. In contrast to the more popular \textit{k-means} algorithm, \textit{k-medoids} chooses data points as centers, named medoids or \textit{exemplars}, that are defined as members of the input set that are the most representative for the assigned cluster. We preferred \textit{k-medoids} to other clustering techniques that find \textit{exemplars}, like affinity propagation \cite{Frey972}, because through the $k$ hyperparameter, we are able to fix the number of clusters to find. Since we are only interested in finding one \textit{exemplar} per dataset, we fix $k = 1$, which is the equivalent of finding the \textit{exemplar} that lies in the middle of the dataset, in its multidimensional vector space.

In a totally unsupervised fashion, it possible to apply this clustering technique to each of the two separate datasets, the clean one $D_C$ and the noisy one $D_T$, that are now perfectly independent, to find a representative (\textit{exemplar}) image for each dataset: one being the \textit{exemplar} for the clean dataset $x_{ex}^{C} \in D_C$, while the other one being the \textit{exemplar} for the noisy one $x_{ex}^{T} \in D_T$.

Going back to the approach defined in Section \ref{measuring-filters-susceptibility}, we can now compute the distance among feature maps activations, for a given convolutional layer $C$, between the \textit{exemplar} for the clean dataset $x_{ex}^{C}$, and the \textit{exemplar} for the noisy dataset $x_{ex}^{T}$. The result, differently from the \textit{associative} scenario, is not a distance matrix anymore. Instead, it is a one dimensional vector of scalar values, each representing the distance of each feature map activation at index $j \in \{1, ..., F\}$, of the convolutional layer $C$, between the clean and noisy \textit{exemplar}.

\subsubsection{Ranking} \label{ranking}
Being the computed distance vector a one dimensional array, there is no need to adopt any voting technique like in the \textit{associative} approach anymore. In fact, it is now possible to simply calculate which are the indices that would sort the vector descendingly, so as to come up with the final ranking of the most affected feature maps activations for the convolutional layer $C$.

As for the \textit{associative} case, the produced ranking will be the building block for the next and final step, which consists in fine-tuning only a subset of the convolutional layer filters, corresponding to the most affected ones by noise, as detailed in Section \ref{model-fine-tuning-on-target-dataset}.

\section{Experimental Setting} \label{experimental-setting}

Here, we describe the various datasets, image distortions, network architectures and clustering details used to validate the proposed \textit{transfer learning} technique.

\subsection{Datasets} \label{datasets}
We used two popular image classification datasets: CIFAR-10 and CIFAR-100 \cite{Krizhevsky09}. CIFAR-10 consists of 60000 32x32 pixels colour images in 10 classes, with 6000 images per class. There are 50000 training images and 10000 test images. The training set contains exactly 5000 images from each class, while the test set contains exactly 1000 randomly-selected images from each class. The classes are completely mutually exclusive (e.g. there is no overlap between automobiles and trucks). The CIFAR-100 dataset is just like the CIFAR-10, except it has 100 classes containing 600 images each. There are 500 training images and 100 testing images per class. We split both, CIFAR-10 and CIFAR-100, using an 80-20 ratio: 40000 images for training and 10000 for validation.

\subsection{Distortions} \label{distortions}
We focus on evaluating two important and conflicting types of image distortions: Gaussian blur and Additive White Gaussian Noise (AWGN), each over 3 levels of distortion severity. Gaussian blur, often encountered during image acquisition and compression, represents a distortion that eliminates high frequency discriminative object features like edges and contours, whereas AWGN is commonly used to model additive noise encountered during image acquisition and transmission.

Since we use datasets with the same input resolution, we use identical sets of distortion parameters for each dataset. For AWGN, we use a noise standard deviation $\sigma_g \in \{5, 15, 25\}$ and $\mu_g = 0$. For the Gaussian blur, we instead use a standard deviation $\sigma_b \in \{0.25, 1.25, 2.25\}$. For both CIFAR-10 and CIFAR-100, the size of the blur kernel is set to 4 times the blur standard deviation $\sigma_b$.

\subsection{Network Architectures} \label{network-architectures} 

Due to the larger number of classes to separate from between CIFAR-10 and CIFAR-100 datasets, which accounts for a greater level of complexity in classifying each input image, we used two different network architectures, specifically: a simple convolutional neural network with two pairs of convolutional and max pooling layers, followed by a fully connected layer with a final 10-way softmax layer for CIFAR-10, and a fully-convolutional network that consists of only convolutional layers with a final 100-way softmax layer for CIFAR-100. We adopt the term "pre-trained" or "baseline" network to refer to any network that is trained on undistorted images.

\begin{figure}[t!]
    \centering
    \includegraphics[width=\textwidth]{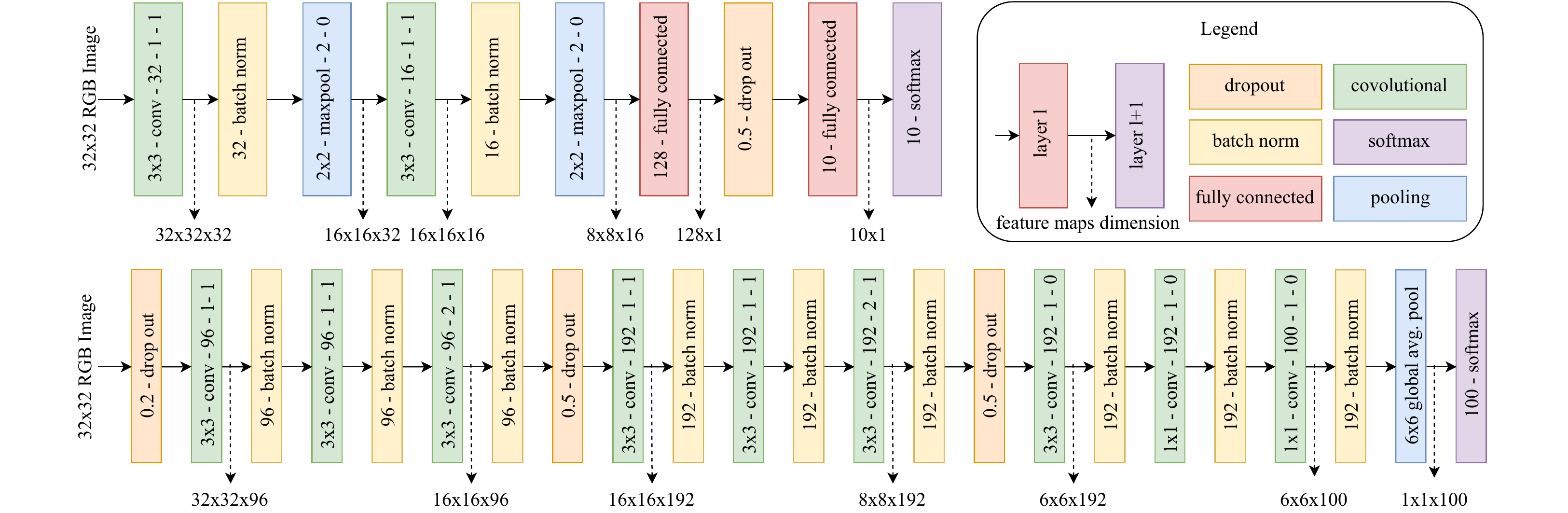}
    \caption{Network architectures for our baseline models. Convolutional layers are parameterized by $k$x$k$-conv-$d$-$s$-$p$, where $k$x$k$ is the spatial extent of the filter, $d$ is the number of output filters in a layer, $s$ represents the filter stride and $p$ indicates the zero-padding added to both sides of the input. Max-pooling layers are parameterized as $k$x$k$-maxpool-$s$-$p$, where $s$ is the spatial stride and $p$ indicates the implicit zero padding to be added on both sides. Batch normalization layers are parameterized by $d$-bn, where $d$ is the number of features in the layer. Dropout layers are parameterized by $pr$-dp, where $pr$ is the dropout probability value. Fully connected linear layers are parameterized by $d$-fc, where $d$ represents the dimensionality of the output space. \textit{Top}: Simple convolutional network for CIFAR-10. \textit{Bottom}: All-Conv Net for CIFAR-100.}
    \label{fig:networks-architectures}
\end{figure}

The architecture for our simple convolutional network, which serves as our baseline model for CIFAR-10 dataset, is very similar to the original \textit{LeNet} one \cite{Lecun98gradient-basedlearning}. The first convolutional layer has 32 filters and kernel size equal to 3, while the second one has 16 filters, with the same kernel size. Both max pooling layers have a kernel size of 2. The fully connected dense layer has 128 neurons, with a dropout probability of 0.5. ReLU nonlinearities are adopted after every batch normalization operation, and as activation functions for the classifier neurons in the fully connected dense layer.

Our version of the fully-convolutional network is based on the All-Conv Net proposed by Springenberg et al. \cite{DBLP:journals/corr/SpringenbergDBR14}, with the addition of batch normalization units after each convolutional layer, and is used as our baseline model for the CIFAR-100 dataset. A summary of both networks architectures can be observed in Figure \ref{fig:networks-architectures}.

For what concerns the training details of the baseline models, we adopted the cross entropy loss for both models, minimized through Adam optimizer, with the standard hyperparameters listed in Adam's paper \cite{DBLP:journals/corr/KingmaB14}. Model fitting was done in an validation-based early stopping setting, adopting a patience hyperparameter of 15 epochs.

\subsection{Clustering} \label{clustering-setting}
The last experimental setting is related to the clustering method, mentioned in Section \ref{non-associative-method}. We decided to experiment different approaches, varying on which features to use to cluster each image and on which distance metric to adopt to compare the data points. For what concerns the features, we decided to cluster either directly on the pixels, where each feature is one of the 32x32 pixels of the image, or on the one-dimensional collapsed version of the baselines' feature maps activations, at the output of the low-level convolutional layers. With respect to the distance metric, only Euclidean distance \cite{anton1993elementary} was considered when features where represented by image pixels, whereas also Hamming distance \cite{warren2012hacker} was tested for the convolutional feature maps case. In case Hamming distance was used, feature maps activations were first converted into binary vectors, setting each element to 1, whenever the corresponding value was greater than 0, whereas set to 0 whenever this condition was not met.

\section{Results and Discussion} \label{results-and-discussion}

\subsection{Noise impact on baseline model} \label{noise-impact-results}

\begin{figure}[t!]
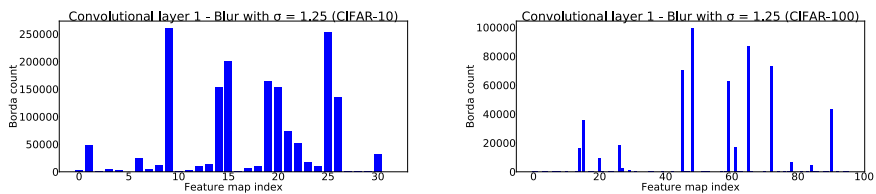

\centering
  \begin{tabular}{@{}cc@{}}
    \includegraphics[width=.49\textwidth]{img/Blur-125-CIFAR10.pdf} &
    \includegraphics[width=.49\textwidth]{img/Blur-125-CIFAR100.pdf}   \\
  \end{tabular}
  \caption{Distortion susceptibility of convolutional filters in the first convolutional layer of both baseline models, when tested on clean and blurred image pairs respectively from CIFAR-10 and CIFAR-100. Similar results can be observed in case of AWGN distortion.}
  \label{fig:bar-plots}
\end{figure}

Before focusing our attention to the effectiveness of the proposed \textit{transfer learning} technique, it is important to demonstrate whether, for a network trained on undistorted images, only some of the convolutional filters in the network are susceptible to noise or blur in the input image. As we can see in Figure \ref{fig:bar-plots}, it is clear how certain convolutional filters are far more susceptible to input distortions than others. Considering for example only the first convolutional layer, in our baseline model trained on CIFAR-10 data, and applying our \textit{associative} method described in Section \ref{noise-impact-analysis}, we can see how some activations always tend to be ranked higher by our voting technique. This demonstrates that their corresponding convolutional filters are far more sensitive to input distortion than others. Restoring the activations of only the filters that are more susceptible to input distortions can reduce the time and computational resources involved in enhancing DNN robustness to such distortions.


Interestingly enough, this set of convolutional filters vulnerable to input distortion, seems to be independent from the type of distortion. Indeed, even though the \textit{Borda counts} are slightly different between the blur and AWGN distortions, the ranking of the most susceptible filters tends to be independent of the type of distortion applied. In fact, if we looked at the rankings produced in the AWGN case compared to the blur one, we could see how the filters that were most affected by noise in the AWGN case, tend to be the same filters in the blurring case, with the exception of a few elements. In fact, considering only the 25\% of the most sensitive feature maps of the first convolutional layer, for both models, 6 out of 8 times there is a match in the selected filters, for the CIFAR-10 case, while a 19 out of 24 ratio for the CIFAR-100 one. This is an interesting result: since the set of filters to fine-tune is shared among the two types of distortion, by simply fine-tuning the baseline models with distorted images from one of the two, the network could end up being robust to the other type of distortion too. This investigation could be subject of future work.

\subsection{Filters fine-tuning on target dataset} \label{filters-finetuning-results}

In this section, we evaluate the proposed approach of fine-tuning baseline models with distortion affected inputs, from the two datasets and architectures mentioned in Section \ref{experimental-setting}. For the reproducibility of the results of the experiments, it is important to note that the classification performance is evaluated independently for each type of distortion. Furthermore, for the baseline trained on CIFAR-10, all shown results are obtained with the fully connected layer unchanged, meaning no fine-tuning was performed on the classifier's weights. We did so, because we wanted to evaluate the impact of our technique on the representation learning capabilities of the network, rather than just measuring the improvement in terms of accuracy. Nonetheless, the reader should be aware that, when performing \textit{transfer learning}, it may also be necessary to retrain a new classifier from scratch, due to differences between source and target tasks. 

\subsubsection{Improving classification accuracy of the baseline model} \label{improving-classification-accuracy}
\begin{figure}[t!]
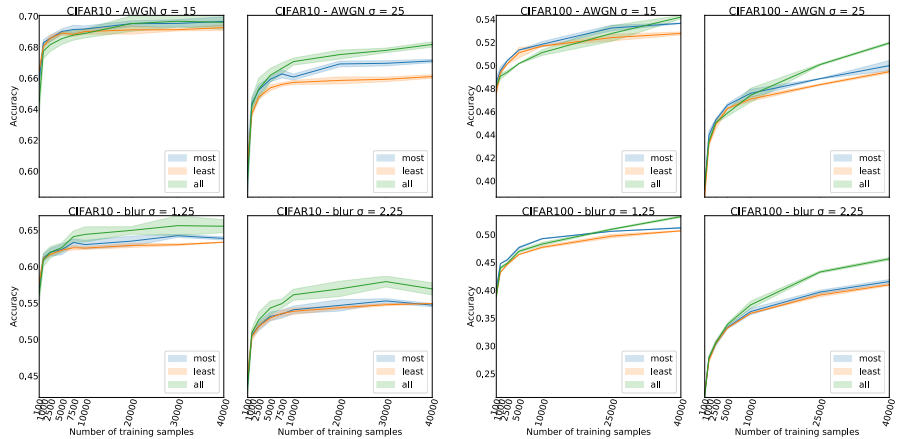

\centering
  \begin{tabular}{@{}cc@{}}
    \includegraphics[width=.49\textwidth]{img/resultsCIFAR102x2results.pdf} &
    \includegraphics[width=.49\textwidth]{img/resultsCIFAR1002x2results.pdf}
  \end{tabular}
  \caption{Fine-tuning effects on the classification performance of noisy inputs, as a function of the number of training distorted images used to perform such fine-tuning. For each plot, three different configurations were considered: (1) \textit{most}: fine-tuning is performed only on the 25\% of the layer convolutional filters most susceptible to image distortion; (2) \textit{least}: fine-tuning is performed only on the 25\% of the layer convolutional filters least susceptible to image distortion; (3) \textit{all}: fine-tuning is performed on all convolutional filters of the layer, independently from their susceptibility to image distortion. For the plots in the first two columns, fine-tuning was done on the baseline model trained on CIFAR-10 undistorted images, correcting the convolutional filters from both convolutional layers of the network. For the plots in the last two columns, fine-tuning was performed on the baseline model trained on CIFAR-100 pristine samples, correcting the convolutional filters only from the first three convolutional layers of the network.}
  \label{fig:retraining-plots}
\end{figure}

One of the major objectives of our proposed \textit{transfer learning} technique is to improve the classification performance of the baseline model when tested on noisy samples. Figure \ref{fig:retraining-plots} summarizes the results of the conducted evaluation of such technique, when only 25\% of the convolutional filters of each of the corresponding convolutional layers are fine-tuned. For the CIFAR-10 baseline, convolutional filters are fine-tuned in both convolutional layers simultaneously, while in the All-Conv Net, trained on CIFAR-100, only early convolutional layers are corrected. Precisely, only the first three convolutional layers. This because, as confirmed by \cite{DBLP:journals/corr/BorkarK17}, the best performance is achieved correcting filters in early convolutional layers of the network. In fact, as we go deeper in the network, accuracy diminishes for correcting a fixed percentage of convolutional filters, which indicates that, as we go deeper in the network, all the convolutional filters become more or less equally susceptible to distortion in the input data.

The results in Figure \ref{fig:retraining-plots}, which plot classification accuracy as a function of the number of noisy training points used to fine-tune the baseline models, clearly demonstrate how, for small training set sizes, fine-tuning only the most affected convolutional filters yields a better classification performance than fine-tuning all the filters in the selected layers, or fine-tuning only the filters that are least susceptible to input distortion, confirming our hypotheses. Instead, for larger dataset sizes, fine-tuning the entire set of filters in the considered layers proves to be more convenient, as expected. 
These conclusions are supported by the graphs at columns 1 and 3 of the tabular representation in Figure \ref{fig:retraining-plots}: independently of the type of noise and network architecture, a moderate level of distortion - as the ones considered in such configurations - is efficiently handled by our technique. In fact, for limited amount of training noisy samples used to fine-tune the baseline networks, the "\textit{most}" configuration is able to outperform the other two, considerably reducing the computational requirements to fine-tune all the convolutional filters in the layer. Instead, when a larger set of noisy samples is available, fine-tuning all the convolutional layers proves to be the most efficient solution.

For what concerns the scenarios that involve the smallest amount of distortion that we studied, which we did not include in the plot for spacing restrictions, other considerations need to be made. In these settings, it is evident how our proposed technique is meaningless, if not even detrimental. For such limited levels of distortion, fine-tuning the baseline models with distorted samples is not necessary, because the \textit{dataset shift} problem, first mentioned in Section \ref{introduction}, is essentially nonexistent, as the the difference between source and target images is fundamentally absent. If we looked at the plots, we would see that the improvement would be in the order of an ideal 2\% margin, which would not justify the need for such \textit{transfer learning} process.

On the other hand, scenarios that involve a substantial level of distortion, as the ones illustrated on the \textit{right} hand-side of Figure \ref{fig:retraining-plots}, present totally different issues. When the input samples are seriously corrupted, two outcomes occur: 1) The "\textit{all}" configuration consistently outperforms the other two configurations (e.g. CIFAR-100 with blur distortion); 2) The "\textit{all}" configuration needs fewer noisy training samples to surpass the "\textit{most}" configuration, with respect to the "central" level of image distortion that we previously discussed (e.g. CIFAR-100 with AWGN distortion). What these configurations results clearly imply is that the higher is the level of distortion, the greater is the number of parameters that need to be corrected to account for such conspicuous corruption. Limiting the number of filters to retrain, would then be too high of a constraint for the baseline model to be able to accordingly adapt to classify so distorted samples. However, it is important to point out that the downsides of the proposed approach on this high level values of distortion is relevant because of the input data resolution. Being CIFAR datasets images so little, the effects of the corruption on the images is substantial. On higher resolution samples (e.g. ImageNet), severely larger corruption intensities would be needed to incur in the drawbacks that we have just presented.

To further improve the classification performance of the baseline model, it is important to note that, in order to maximize the effectiveness of the proposed technique, fine-tuning a subset of the most affected filters from multiple layers proved to be more successful than fine-tuning only a single layer of the network at a time. This result proves to be intuitive, as a larger amount of parameters is fine-tuned, enabling a potentially greater level of correction capabilities on the network. Nonetheless, the number of parameters is still limited, with respect to fine-tuning all the convolutional filters of the considered layers, causing our proposed technique to outperform the usual fine-tuning of the entire layer parameters, in the aforementioned circumstances, while also limiting the computational requirements to perform the fine-tuning. 

In conclusion, it is also relevant to point out that, as mentioned at the beginning of the section, we decided to fine-tune only one fourth of the convolutional filters present in each convolutional layer. This decision was not based on a thorough assessment of how many filters to fine-tune per layer and we leave this task as a subject for future work. A starting point could be to fine-tune only the set of convolutional filters whose \textit{EMD} metric is above a certain threshold. By doing so, we would make sure to fine-tune only the filters who account for the greatest degradation in performance, and not include also those who happen to be in the considered 25\% but do not vary as much as the other ones at the top of the ranking. Nonetheless, it is clear how our convolutional filters ranking methodology, detailed in Section \ref{noise-impact-analysis}, proves to always be beneficial. In fact, fine-tuning the top 25\% of the convolutional filters most susceptible to input data distortion - according to our ranking - is consistently better than retraining the bottom 25\%. Evidence on this remark can be observed in every configuration depicted in Figure \ref{fig:retraining-plots}.

\subsubsection{Reducing the covariate shift effects} \label{reducing-covariate-shift-effects}
\begin{figure}[t!]
    \centering
    \includegraphics[width=\textwidth]{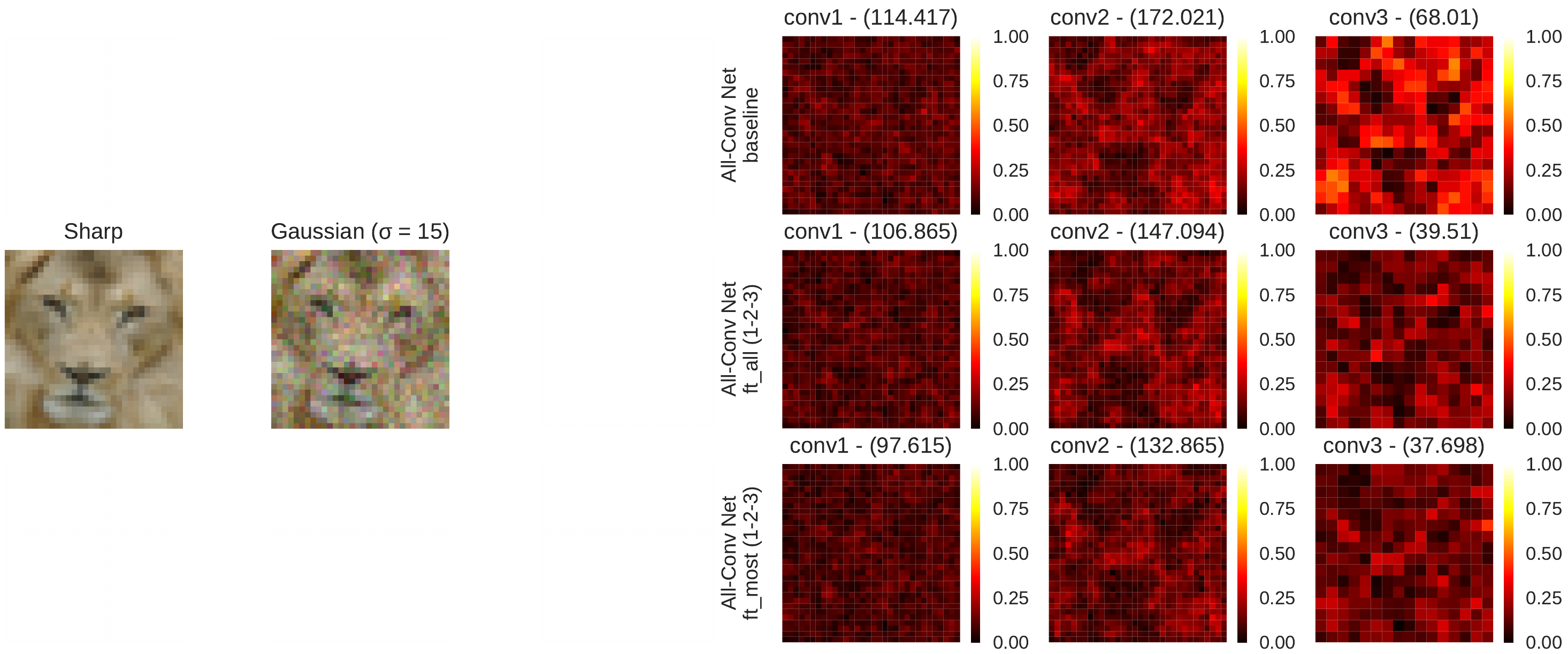}
    \caption{Disparity between corresponding layer activations on sharp and noisy versions of an example image. Each heat-map represents the Hamming distance between binarized feature vectors (i.e., if each channel is positive or zero) at corresponding locations in the sharp and Gaussian distorted inputs. We visualize these distance maps for the first three convolutional layers in the All-Conv Net architecture, comparing three different models: (a) Baseline, trained on undistorted images; (b) Network where all the filters in the first three convolutional layers where fine-tuned with 5000 Gaussian distorted training images from CIFAR-100; (c) Network where only the top 25\% of the convolutional filters most susceptible to Gaussian distortion in the first three convolutional layers where fine-tuned with 5000 Gaussian distorted training images from CIFAR-100. The numbers between the round brackets indicate the element-wise sum of each element in the corresponding heat-map: the higher the number, the larger is the disparity between the matching layer activations on the sharp and noisy version of the example image. We see that model where only the filters most susceptible to Gaussian distortion where fine-tuned produces feature activations that are relatively invariant to the presence of Gaussian noise in the input image.}
    \label{fig:hamming-distance}
\end{figure}

As the last intrinsic result of our \textit{transfer learning} technique, we recall that one of the fundamental objectives of our method was to overcome the problem of \textit{covariate shift}, first mentioned in Section \ref{problem-definition}. This phenomenon was said to be responsible for the degradation in performance of our baseline models when tested on distorted images. The reliable performance of the fine-tuned models to several noise intensities suggests that the fine-tuned networks learned to be invariant to such noises. Inspired by \cite{DBLP:journals/corr/VasiljevicCS16}, we replicated their empirical evaluation methodology to assess such feature invariance to input distortion. In detail, we look at the similarity in activations of different layers of both the baseline and the fine-tuned All-Conv Net model, when sharp and noisy versions of the same image are given as input to the network. Specifically, we consider the feature maps at the output of the first three convolutional layers in the All-Conv networks, being them the only convolutional layers that were subject to our fine-tuning approach. We convert the feature vector at every location into a binary string representing whether each feature channel had a positive or zero response. In Figure \ref{fig:hamming-distance}, we visualize Hamming distances between corresponding binary strings produced from a sharp and AWGN distorted versions of the same example image, for the baseline All-Conv model, and the models fine-tuned on images affected by AWGN, both when the fine-tuning was done on all convolutional filters per layer, or only the most affected ones. What we can assess from the heatmaps in the figure is that the baseline model produces different activations on the sharp and noisy inputs, at all layers. In contrast, the fine-tuned models are able to achieve a reasonable amount of noise invariance, with low distances between sharp and distorted activations in all three fine-tuned layers, with the model fine-tuned only on the most affected convolutional filters being slightly more invariant to image distortion than the one where all the filters were fine-tuned. This result is in line with our hypotheses and positively exhibits the reasoning behind our assumptions.

\subsection{Non associative ranking} \label{non-associative-ranking}

To complete our discussion about the evaluation of the proposed methodology, we evaluate the \textit{non associative} technique proposed in Section \ref{non-associative-method}. As previously stated, this method is meant to provide a solution to the case in which the \textit{(clean, noisy)} image pair, for each training image, is not available. We presented a way to overcome this problem, relying on the so called \textit{exemplars}.

Table \ref{table:1} shows, for the first three convolutional layers in the All-Conv Net baseline model, trained on CIFAR-100, how many filters identified by the different configurations of our \textit{non associative} approach actually match the set of filters identified by the \textit{associative} technique.

As we can see from the table, the numbers of matching filters is definitely promising, with the configuration that uses image pixels being the one that - overall - was able to find the ranking closest to the \textit{associative} one. It is important to note that, even though such lists are produced as rankings in the first place, the actual ordering is not relevant, because each of them simply represents a set of convolutional filters that need to be fine-tuned. Because of this, the fact that a given index value comes later in the ranking from the \textit{non associative} approach than the \textit{associative} one, it is not going to have an impact on the fine-tuning performance. Therefore, the highest the number of matching indices, the better will be the fine-tuning. 
This result guarantees that fine-tuning the convolutional filters indicated by this technique will actually achieve approximately the same performance of our \textit{associative} approach, assuring the applicability of our methodology to every empirical setting, \textit{associative} or not.

\begin{table}[t!] 
\centering
\begin{tabular}{m{2cm}|m{1.5cm}|m{1.5cm}|m{1.5cm}|m{1.5cm}|m{1.5cm}|m{1.5cm}|}
\cline{2-7}
  & \multicolumn{2}{c|}{\textbf{Layer 1}} & \multicolumn{2}{c|}{\textbf{Layer 2}} & \multicolumn{2}{c|}{\textbf{Layer 3}} \\
 \hline
 \multicolumn{1}{|c|}{\textbf{Distance}} & \hfil Pixels & \hfil FTM & \hfil Pixels & \hfil FTM & \hfil Pixels & \hfil FTM \\
 \hline
 \multicolumn{1}{|c|}{Euclidean} & \hfil 18/24 & \hfil 14/24 & \hfil 13/24 & \hfil 13/24 & \hfil 19/24 & \hfil 19/24 \\
 \hline
 \multicolumn{1}{|c|}{Hamming} & \hfil - & \hfil 13/24 & \hfil - & \hfil 11/24 & \hfil - & \hfil 17/24 \\
 \hline
\end{tabular}
\vspace{1.5ex}
\caption{Comparison of the number of matching convolutional filters, per layer, between the \textit{associative} ranking and several configurations of the \textit{non associative} ones. All the \textit{non associative} rankings, and the \textit{associative} one used as "ground truth", are based on the All-Conv Net baseline model, trained on undistorted images from CIFAR-100. All three convolutional layers have 96 filters each, so the top 25\% of each layer only considers 24 filters. The noisy images that are used to perform the comparison between clean and noisy activations were perturbed with AWGN with distortion severity $\sigma = 15$ (comparable results are obtained when blurring distortion is in place). \textit{Pixels} indicates image pixels were used as features for the clustering method, whereas \textit{FTM} when the collapsed version of the baseline feature maps activations - at the output of corresponding convolutional layer - were adopted.}
\label{table:1}
\end{table}

\section{Conclusion} \label{conclusion}
Deep neural networks trained on uncorrupted images perform poorly when tested on distorted images affected by image blur or additive white Gaussian noise. Evaluating the effect of Gaussian blur and AWGN on the activations of convolutional filters trained on undistorted images, we observe that some of the filters in each convolutional layer of the CNN are far more susceptible to input distortions than others, and prove that this set of filters is nearly independent from the distortion applied to the input data. 

We propose a novel \textit{transfer learning} technique to assess the susceptibility of convolutional filters to input data distortion and use this procedure to identify the filters that contribute the most to the drop in classification performance that occurs when a DNN, trained on undistorted images, is tested on perturbed ones. We demonstrate how our assessment is also applicable to situations in which the correspondence between clean and noisy versions of the same image is not available, providing a solution to scenarios that related works in the literature would fail to assist.

We design a new way to perform \textit{transfer learning}, moving on from the usual fine-tuning of all convolutional filters of selected layers of the DNN, improving the robustness of the network against image distortions by simply fine-tuning only the most distortion-susceptible convolutional filters of the model, while leaving the rest of the pre-trained (on undistorted images) filters in the network unchanged. The resultant DNN models outperform DNNs fine-tuned with the usual fine-tuning approach, when labeled data in the noisy domain is limited, and achieves a comparable performance when training data in this setting is vastly available. Additionally, our proposed technique requires training a significantly lower number of parameters than the conventional fine-tuning approach - that would be computationally unfeasible on very large networks - or than ad hoc correction units that are added at the output of the most distortion susceptible filters in each convolutional layer, that would exponentially increase the number of parameters to train. Such results are achieved with faster convergence in training while still accomplishing the task of learning features invariant to distorted input data.

%
%
%
\bibliographystyle{splncs04}
\bibliography{bibliography}

\end{document}